# PI(t)D(t) Control and Motion Profiling for Omnidirectional Mobile Robots

Michael L. Zeng

*Abstract*— Recently, a trend is emerging toward human-servicing autonomous mobile robots, with diverse applications including delivery of supplies in hospitals, hotels, or labs where personnel are scarce, or reacting to indoor emergencies. However, existing autonomous mobile robot (AMR) motion is slow and inefficient—a foundational barrier to proliferation of human-servicing applications. This research has developed a motion control architecture that demonstrates the potential of several algorithms for increasing speed and efficiency. These include a novel PI(t)D(t) controller that sets integral and derivative gains as functions of time, and motion-profiling applied for holonomic motion. Resulting performance indicates potential for faster, more efficient AMRs, that maintain high levels of accuracy and repeatability. The hope is that this research can serve as a proof of concept for faster motion-control, to remove a key barrier to further use of human-servicing mobile robots.

*Keywords—robotics, mobile robotics, motion control, motion profiling*

## I. INTRODUCTION

State-of-the-art autonomous mobile robots use an array of sensors such as lidar, depth cameras, and odometry to achieve localization and real-time detection of obstacles, and are capable of generating and following smooth paths and avoiding obstacles in real time [1][2]. Today, a vast majority of autonomous mobile robots operate in warehouses and factories, typically, repetitively transporting heavy loads, where speed of the robots is not as much of a concern. Recently, there has been a trend toward applying autonomous mobile robot technology in human-servicing applications [3]. Moxi, developed by Diligent Robotics, is a good example. Named one of the top 100 inventions of 2019 by TIME, Moxi is a robot that delivers light supplies in hospitals, so that nurses and staff, who are in short supply in many places, can focus their efforts on patient care [4]. Numerous other startups and research groups are also emerging with service autonomous mobile robot technology. Other robotic couriers also operate in hospitals [5][6], robots have been studied in caring for the elderly [4], numerous mobile robots have been developed to interact with humans and/or deliver supplies in hotels or restaurants, and a group is researching mobile robots to autonomously perform chemistry experiments in a lab [7].

However, there is still a fundamental barrier to expanding use of autonomous mobile robots in human-servicing applications: speed and efficiency of motion. Navigation of mobile robots requires high robustness in order to function in a variety of environments without causing safety concerns. Therefore, existing autonomous mobile robots travel at low speeds and low accelerations, with movement patterns that take unnecessary amounts of time (i.e., inefficient path-planning, or stopping unnecessarily and losing momentum). With the example of Diligent Robotics' Moxi robot, Moxi travels at an estimated 17 in/s and takes about 5 seconds for a 360-degree rotation (these estimates are based on online video-reference material), while a human walks at an estimated 55 in/s. These speeds mean the robot is not nearly as effective as a human and has much room for improvement. In addition, Moxi navigates on a mobile robot base developed by Fetch Robotics, the leader in the autonomous mobile robot industry [8] and represents state-of-the-art mobile robot technology that is available today. While autonomous mobile robots move so slowly, the proliferation of their use alongside humans in human-centric environments is not very efficient or practical.

Mobile robots can generally be classified as either holonomic (if its wheelbase has three degrees of freedom), or non-holonomic (its wheelbase has two or fewer degrees of freedom). A holonomic wheelbase was chosen for this study, because it is always desirable for a mobile robot to have full control of the three degrees of freedom (although, at a cost of vibration and energy efficiency), which results is unrestricted motion in all directions along the plane of the ground [9]. Mecanum wheels are used most often for holonomic mobile robots for their reliability and simplicity.

State-of-the-art, traditional control systems must use a combination of methods to achieve navigation. Robust sensors first must provide localization and mapping data. An algorithm must then calculate a path for the robot to follow. Finally, the robot must have some method(s) of following this path. In regard to the path-following task, two common methods include PID control and motion profiling.

Proportional-integral-derivative (PID) control is currently the most widely used control method in industry and robotics and performs robustly on a variety of systems [10]. For robust control, PID gains must first be tuned to the system. There has been significant progress made in efficiently and effectively tuning PID gains. The Zeigler-Nichols tuning method is frequently used. Onat developed graphical methods for tuning PID gains [11]. To solve the problem of needing expert knowledge on the system behavior in order to manually tune the controller, there have also been many automatic PID tuning methods that can maintain robust control with unknown or time-varying system behaviors. Tang proposed an optimal PID controller in which the PID gains are self-tuned using fuzzy logic [12]. Mitsukura has developed a self-tuning PID controller using a genetic algorithm and generalized minimum variance control laws, which just requires some suitable user-specified parameters [13]. Papadopoulos developed an automatic tuning PID controller using the magnitude optimum criterion for a typical single-input single-output plant [14]. However, traditional PID control methods fail to achieve optimal speed and control performances at the same time, especially for mobile robots, which are typically mechanically capable of high accelerations; PID is a linear controller, while mobile robots are non-linear systems [15], [16], [17]. The result is that a PID controller sacrifices robots' speed in order to minimize overshoot and oscillations.

Motion profiling is another frequently used method of motion control. By knowing the mechanical constraints of the system, motion profiles predict kinematic information (such as acceleration and velocity) as a function of time for a particular path. These predicted kinematic information are then used to influence the robot's real behavior. Trapezoidal motion profiles are a common method of motion profiling; this approach assumes the magnitude of the robot's acceleration is binary—either zero or a predefined constant. Trapezoidal profiles tend to cause overshoot, vibrations, and require more time to reach a desired position [18]. S-curve motion profiles perform better and are also common; in this approach, the magnitude of the robot's jerk is binary—either zero or a pre-defined constant [19], [20]; this more closely models the behavior of a real robot. Motion profiling itself is a feedforward control method, meaning it does not take any sensor feedback and doesn't react to random error, and therefore doesn't perform well on its own. It is primarily beneficial for reliability and repeatability. In addition, motion profiling has seldom been researched for holonomic mobile robots.

The purpose of this research is to propose a motion control framework that improves speed and efficiency of AMRs to enable proliferation of applications. The primary metrics are to reach higher speeds and efficiency compared to state-of-the-art methods, while maintaining industry-standard safety.

## II. METHOD

*2.1 General Architecture*

The proposed architecture enables full functionality of an autonomous mobile robot. It contains the following components: sensors, such as encoders, lidar, or cameras, first are used to localize the robot and generate a map of surroundings. An A* algorithm or similar algorithm can be used to generate the most efficient path from the robot's current position to a target position while avoiding obstacles [21]. Finally, the robot follows the path using two main methods: 1) a novel, highly optimized controller for mobile robots called the PI(t)D(t) controller, combined with a pure pursuit algorithm, and 2) holonomic motion profiling.

This research focuses on the path-following step.

The holonomic motion profiling is a form of feedforward control and is important for increasing consistency between trials, while the PI(t)D(t) controller with pure pursuit algorithm is a form of feedback control and is necessary to correct for deviations from the target path. The outputs of the two path-following methods are superimposed to govern the robot's motion. The overall control system flow diagram is indicated below:

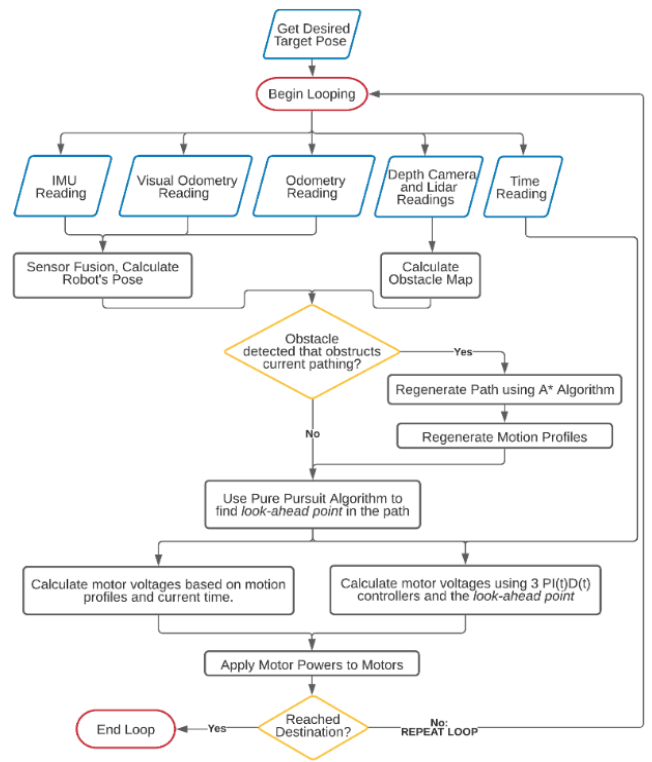

Fig. 1. Full Control System Flow Chart.

*2.2 Holonomic Motion Profiling*

Motion profiling utilizes the robot's measured constraints and a path (a series of coordinates containing X, Y, and heading) to project the robot's kinematics (velocity, angular velocity, acceleration, and angular acceleration) as a function of time when following the path [22]. This serves a number of purposes. Firstly, knowing the projected velocity and acceleration of the robot at a point in time can be used to influence the robot's motion. Secondly, projecting the behavior of the robot enables a prediction of the time necessary to follow a path. This is generally useful information for practical applications but is also vital to the functionality of the PI(t)D(t) controller, which will be discussed later.

The robot's constraints must first be measured. These constraints include:

- $kv_x$, $kv_y$, $kv_h$ are the slopes of linear velocity vs input voltage curves for x, y, and heading, respectively. These were measured using a quasi-static voltage ramp-up test.
- $ka_x$, $ka_y$, $ka_h$ are the slopes of linear acceleration vs input voltage curves for x, y, and heading, respectively. These values were tuned manually; in the tuning process, the robot accelerates, then moves at a constant velocity, then decelerates, in a straight line, using just feed-forward control. The $ka$ values were then tuned such that the real velocities of the robot matched the target velocities as closely as possible.
- $v_x^{max}$, $v_y^{max}$, $v_h^{max}$ are the maximum linear velocities of the robot for each degree of freedom (when the other two degrees of freedom are fixed). These were measured by inputting the maximum voltage to the robot's motors and measuring the velocity when it stabilizes. Alternatively, the maximum velocities can be decreased if the system is unsafe at the measured $v^{max}$.
- $a_x^{max}$, $a_y^{max}$, $a_h^{max}$ are the maximum linear accelerations of the robot for each degree of freedom (when the other two degrees of freedom are fixed). These were measured by rapidly accelerating the robot from rest and measuring the resulting acceleration.
- $j_x^{max}$, $j_y^{max}$, $j_h^{max}$ are the maximum linear jerks of the robot for each degree of freedom (when the other two degrees of freedom are fixed). These values were estimated and manually tuned.

When creating the motion profiles, the robot first designates a limited power for rotation and translation to realistically model the robot. Because the robot has to sustain simultaneous motion in all three degrees of freedom, the effective $v_x^{max}$, $v_y^{max}$, $v_h^{max}$ are different from the values that were measured when testing each degree of freedom independently. This difference is proportional to the ratio of power needed for translational and rotational motion (i.e., if a path contains significant rotation, less power can be allocated the translation and the maximum linear velocity will be reduced). (Note that this initial step is only necessary if the robot is expected travel at or near its maximum input power at any time). For ease of calculation, it is assumed that the acceleration and jerk take negligible time compared to the time spent at a stable "travel velocity" and "angular travel velocity". We can denote the robot's travel velocity as $v$, and the robot's angular travel velocity as $v_{ang}$. Knowing the path's total length $\Delta x$ and total angular displacement $\Delta h$, we know that $\frac{\Delta x}{v} = \frac{\Delta h}{v_{ang}}$, because the time taken for translation should equal the time taken for rotation (as they would happen synchronously). Therefore, we can find the ratio $\frac{v}{v_{ang}} = \frac{\Delta x}{\Delta h}$. Using $kv_x$, $kv_y$, and $kv_h$, which determine the linear relationship between velocity and motor voltage, we can determine the ratio of voltage that should be alotted to translation and rotation. The ratio of voltage can be used to determine a new $v_{linear}^{max}$ and $v_h^{max}$ for this particular path. This step is the primary difference between the holonomic motion-profiling algorithm developed in this research and non-holonomic algorithms.

The motion profiles for linear velocity, angular velocity, linear acceleration, and angular acceleration, are then calculated separately, similarly to most existing motion profiling methods. A seven-part motion profile is used for each; to demonstrate what this looks like, a real motion profile for linear acceleration can be seen below with the corresponding position vs time plot:

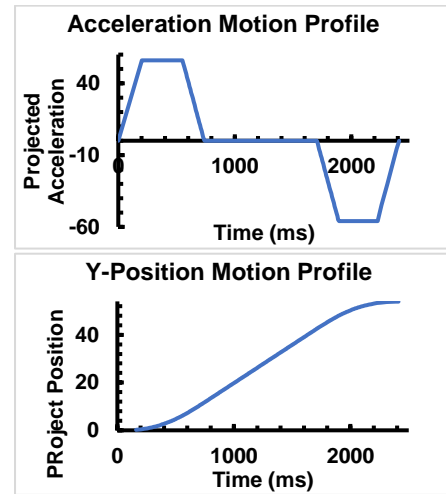

Fig. 2. Example Acceleration and Position Motion Profiles.

To create this seven-part plot, constant-jerk kinematic equations are used. The first three sections bring the hypothetical robot up to maximum velocity, and the last three sections return the robot to zero velocity. In the middle section, the robot is at zero acceleration at maximum velocity. In the first, third, fifth, and seventh sections, the robot is at its maximum jerk. In the second and sixth sections, the robot is at its maximum acceleration. By solving series of kinematic equations, the robot can determine the precise timing of each section of the motion profile, to ensure that the kinematics enable the robot to reach its precise destination as efficiently as possible.

It is worth noting that the motion-profiling calculations assume the robot will reach maximum velocity. However, if the kinematic equations demonstrate that the path is too short, such that the robot is unable to reach maximum velocity before reaching its target, motion profiling is simply not used. Instead, the PI(t)D(t) controller with pure pursuit algorithm is used alone. When the path is this short, the task resembles correcting for large errors rather than following a long path, so feedback control is more fit for the situation.

### 2.3 PI(t)D(t) Controller with Pure Pursuit

The PI(t)D(t) controller and pure pursuit algorithm work together to both follow the path and correct for error. The pure pursuit algorithm first looks for a point on the path to "follow", then the PI(t)D(t) controller maneuvers the robot toward that point. The PI(t)D(t) controller is a novel, highly optimized controller for mobile robots.

The pure pursuit algorithm is a well-established algorithm used in path-tracking [23][24]. The robot has a "look-ahead distance", $d$. The robot will look for a point ahead on the path that is a distance $d$ away; $d$ is typically a hyper-parameter defined by the user. A larger look-ahead distance will result in smoother motion but more deviation from the path. A smaller look-ahead distance requires the robot to follow the path more strictly. For the proposed control system, the look-ahead distance is tuned based on wheel-slippage (this will be further discussed below) but is generally in a middle-ground.

The PI(t)D(t) controller functions much like standard PID controllers. Three of these controllers are used—one governs x-axis position, one governs y-axis position, and one governs robot heading. Each controller takes the plant error—the error between the robot's current pose and the look-ahead point—as input, and outputs a 1D vector between 0.0 and 1.0 in the corresponding degree of freedom (x, y, or heading), where 0.0 denotes zero velocity, and 1.0 denotes maximum velocity.

Like a PID controller, the PI(t)D(t) has a proportional, integral, and derivative output, however, each is calculated differently. The proportional output is calculated as $k_p$ (proportional gain) times error, denoted by $e$. However, performance was improved by limiting the acceleration on the robot to improve traction. Acceleration is controlled by two hyperparameters; the "start-power" $u_0$, and "ramp-up" $a$. The start-power is the greatest velocity the wheels can be driven at to accelerate the robot from rest without wheel slippage. The ramp-up is the robot's maximum linear acceleration. The proportional output begins at $u_0$, and ramps the speed up, while gradually decreasing as the robot nears the target. A max function is used to ensure the ramp-up doesn't push the proportional output over 100% speed. The proportional output $u_p$ is calculated as follows:

$$u_p = k_p * \max(\{u_0 + a(1-e), 1.0\}) * e \quad (1)$$

The integral output depends on a few more variables, which we shall first define. $t$ is the current time elapsed since the controller began controlling the robot. $T$ is the estimated time for the entire desired motion, which was calculated with the motion profiles. $k_i$ is the integral gain. To calculate the integral output, first we take the integral of $e$. This integral is then square-rooted to increase the relative significance of small errors, which are hard to correct for mobile robots due to friction. This quantity is multiplied by the ratio of $(\frac{t}{T} + 1)$; this enables the integral output to increase as time passes and the robot nears the end of the path. The reason for this is that, at the end of the path, the robot has very little momentum, and the proportional output is low, so small steady-state errors are hard to correct for. Finally, this quantity is multiplied by $k_i$, to calculate the final integral output $u_i$:

$$u_i = k_i * \sqrt{\int_0^t e} * (\frac{t}{T} + 1) \quad (2)$$

To calculate the derivative output of the PI(t)D(t) controller, the derivative of the error is first calculated. This derivative is then divided by the "time-factor", which is calculated as the fourth power of the quantity: $t$ divided by $T$ plus 1. The time-factor has the effect of exponentially decreasing the derivative output as the robot reaches the target; this enables extremely high deceleration when the robot has a lot of momentum, and lesser deceleration when the robot is near the end of the path. This is critical to achieving a short stopping distance while allowing the robot to maintain high travel speeds. Finally, this quantity is multiplied by $k_d$ (the derivative gain), to calculate the final derivative output $u_d$:

$$u_d = k_d * \frac{\frac{de}{dt}}{(\frac{t}{T} + 1)^4} \quad (3)$$

The control law of the entire PI(t)D(t) controller is indicated as below (the sum of the proportional, integral, and derivative outputs):

$$\begin{aligned} u_{net} &= u_p + u_i + u_d \\ &= k_p * \max(\{u_0 + a(1-e), 1\}) * e \\ &+ k_i * \sqrt{\int_0^t e * \left(\frac{t}{T} + 1\right)} \\ &+ k_d * \frac{\frac{de}{dt}}{(\frac{t}{T} + 1)^4} \end{aligned} \quad (4)$$

Because three PI(t)D(t) controllers are used at once to control the robot's three degrees of freedom, one or more of these may have the target position equal to the robot's starting position, or very close (in other words, the starting error of a controller is 0). Take for example, moving in a straight line in the y-axis; this requires the x-axis and heading to remain unchanged. In addition, the difference between the target and starting state (in other words, the starting error of the controllers) may be different for different paths; the difference for heading can be anything between 0 and 180 degrees depending on the desired path, and the look-ahead distance (which is constant), which is the difference in XY position between the target and starting state when the pure pursuit controller is in use, can be divided between the X and Y axes. Therefore, it makes sense for the error of the controller to be calculated as a percentage: $\frac{current\ error}{starting\ error}$. When the starting error is close to zero, the percentage error is close to infinite, which results in erratic robot behavior. To stabilize the behavior, a function was created to scale the starting error up when it is below 8.5 inches, which is indicated below ($y$ is the scaled starting error; $x$ is the raw starting error):

$$\begin{aligned} y &= x + \frac{5}{0.6(x + 0.9) + 1} \quad \{x < 8.5\}, \\ y &= x + 0.746 \quad \{x > 8.5\} \end{aligned} \quad (5)$$

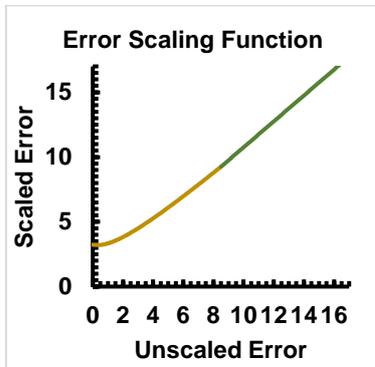

Fig. 3.    Error-scaling function for PI(t)D(t) controller.

Overall, the PI(t)D(t) controller enables significantly faster travel speeds for the robot, mainly due to more effective deceleration. The controller also handles acceleration and steady-state errors more effectively than a PID controller.

## III. EVALUATION AND DISCUSSION

*3.1 Mechanical Prototype*

To test the effectiveness of the motion control system, a holonomic mobile robot platform was developed. The robot is approximately 0.45x0.45x15 meters in size, and approximately six kilograms. The robot utilizes 12V brushed DC motors and a 12V NiMH battery and has a RK3328 Quad-core ARM Cortex-A53 processor with 8 GB of memory. These components were chosen to develop a cost-effective testing platform that could run different motion control frameworks and allow for direct comparison.

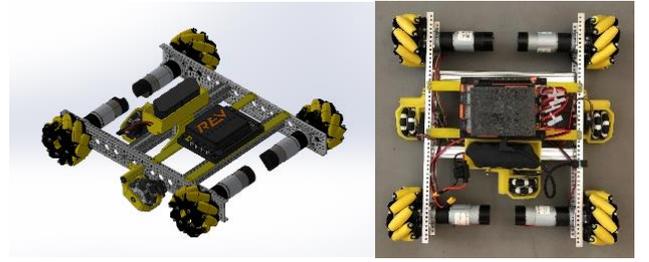

Fig. 4.    Robot Prototype.

The sensors that the robot has onboard include three encoders and an inertial measurement unit (IMU). The three encoders are each attached to an omni-wheel to create three "odometry modules"; two odometry modules face the robot's y-axis and are positioned at the right and left side of the robot, while one odometry module faces the robot's x-axis and is positioned near the back of the robot. Using some trigonometry, these three cost-effective odometry modules can be used to achieve accurate localization.

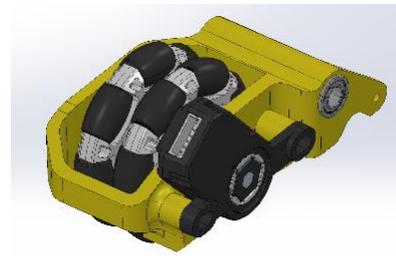

Fig. 5.    Single Odometry Module.

*3.2 Mecanum Locomotion*

The motion-control frameworks described in this research can be applied to any holonomic mobile robots. However, a mecanum drive was chosen for testing, for their simplicity and popularity in holonomic autonomous mobile robots.

A simple thrust-vectoring algorithm was developed to convert the X, Y, and rotation velocity vectors outputted by the PI(t)D(t) controller and motion profiles into voltages for all 4 motors that will enable the specified combination of X movement, Y movement, and rotation. Similar thrust-vectoring algorithms can be used for other types of holonomic robots.

*3.3 PI(t)D(t) Controller Evaluation*

The effectiveness of the PI(t)D(t) controller was first evaluated independently from the rest of the motion control system. A manually-tuned PID controller was chosen as a baseline for comparison for the PI(t)D(t) controller [10]. Two main evaluations were conducted. Firstly, the controllers moved the robot from a starting location to a setpoint, a certain distance away. Secondly, the controllers were integrated with a Pure Pursuit algorithm to allow the robot to follow a simple straight-line path, in order to test the path-following speed of each controller. In each evaluation, the setpoint is considered reached if the robot stops within 1 inch or less of error, and with a negligible velocity.

In this first evaluation, the robot began 1 ft., 2 ft., 3 ft., 4 ft., 6 ft., 9 ft. from the setpoint. Two trials were run for each distance, once using the PID controller, and a second trial using the PI(t)D(t) controller. The error in the system (which is measured as the distance between the robot and target) is plotted against time for each trial:

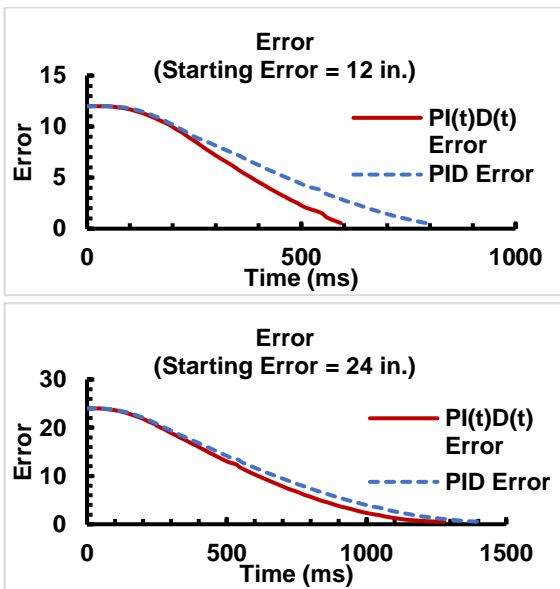

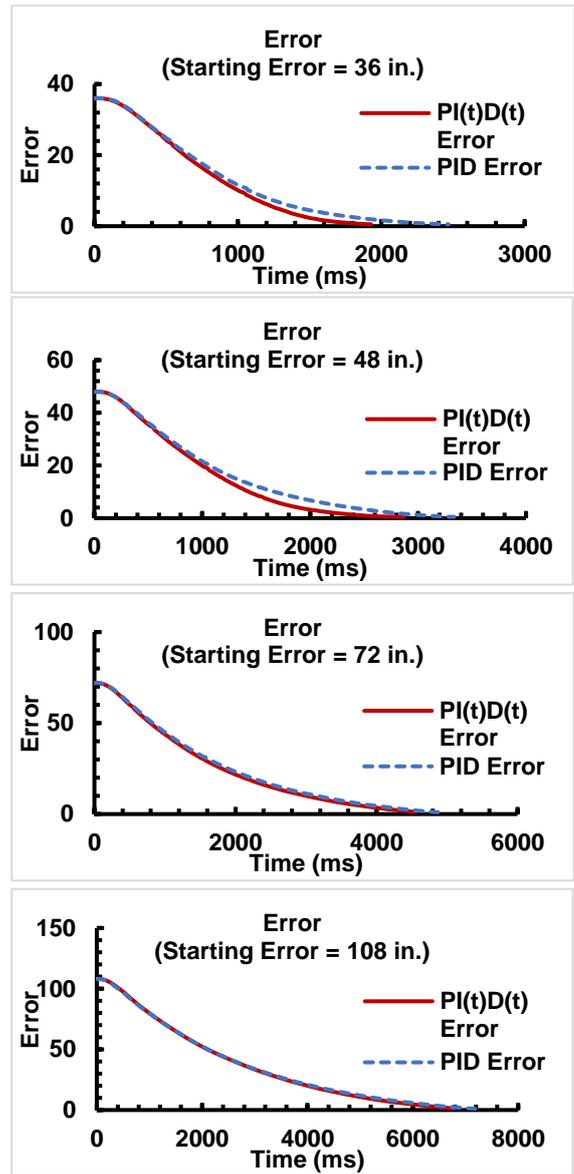

Fig. 6. Error plots for each controller test trial.

As shown in Fig. 6, the PI(t)D(t) is always able to reach the setpoint in less time compared to PID. Both systems experience none or negligible overshoots. The amount of improvement between PI(t)D(t) and PID for each trial is plotted below, where the y-axis denotes the percentage less time taken to reach the setpoint:

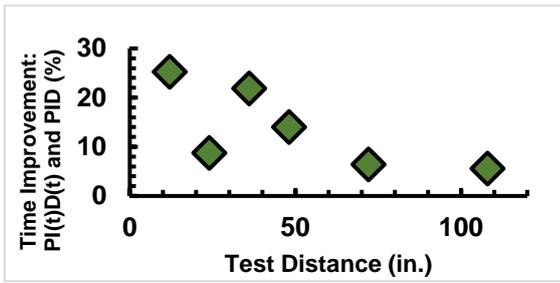

Fig. 7.  PI(t)D(t) Performance Improvement (in percentage less time taken) at Different Distances.

The general trend is that PI(t)D(t) significantly outperforms PID when the test distance is small, and the performance improvement settles at about 5% as the test distance increases.

Because PI(t)D(t) has much more robust acceleration and deceleration, the tuned gains for PI(t)D(t) enable it to travel at high maximum speeds, at which a PID controller would face oscillations near the target.

In the second evaluation, the robot follows a nine-foot-long straight path for ten trials, using a pure pursuit algorithm combined with a either PID or PI(t)D(t) controller. Five look-ahead-distances are tested: 6 in., 8 in., 12 in., 16 in., 24 in., for two trials each (one with PI(t)D(t) controller, and one with PID controller). The purpose of this test is to compare PI(t)D(t) with PID controllers when following a path, and to determine the impact of the look-ahead-distance. The error in the system is plotted against time for each trial (only two trials are shown below):

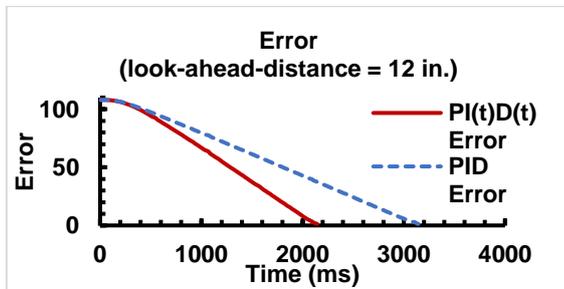

Fig. 8.  Error vs Time plot for path-following controller test.

During the path-following trials, the pure pursuit algorithm with PI(t)D(t) performs significantly better, reaching near the setpoint in approximately $\frac{2}{3}$ of the time taken by the PID controller. This is likely because, after accelerating up to speed and until reaching within the look-ahead distance to the end of the path, the pure pursuit controller allows the robot to go continuously at the robot's maximum velocity. The PI(t)D(t) enables a much higher maximum velocity, and the performance difference between PI(t)D(t) and PID increases over time.

In addition, the optimal look-ahead distance was determined to be roughly 12 inches for the PI(t)D(t) system and, 6 inches for the PID system (for the PI(t)D(t) system, a 6-inch look-ahead-distance resulted in instability and wheel slippage at the end of the path). However, the time difference between 6- and 12-inch look-ahead distances was insignificant; therefore, a look-ahead distance of 12 inches was used during the entire control system evaluations (described below).

*3.4 Control System Evaluation*

To test the entire control systems, 60 total trials were run. 3 different test scenarios with varying complexity were used, and each scenario was tested for 10 trials using PI(t)D(t) controller with Pure Pursuit with Motion Profiling (this will be called "PI(t)D(t) system" for short), and 10 trials using just PID controller with Pure Pursuit (this will be called "PID system" for short). The 3 test scenarios are as follows (the arrows denote the robot's target heading at various points in the path, the red dot denotes the target pose, and the robot icon denotes the starting pose):

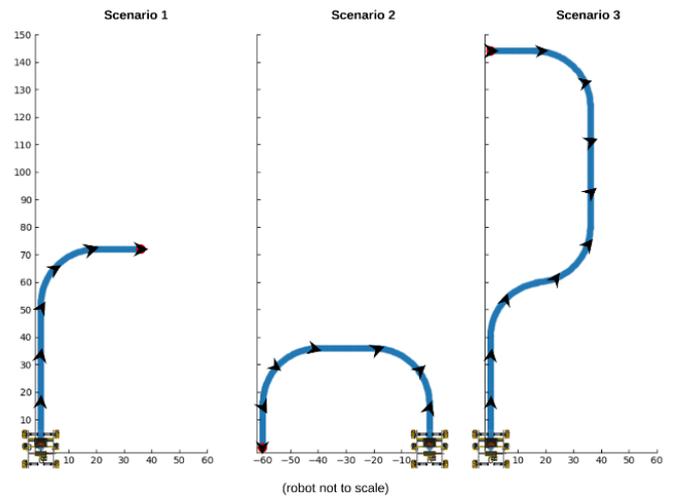

Fig. 9.  3 test scenarios.

Only the robot's X and Y position were considered in this evaluation, as rotation is quicker than translation, and therefore heading does not affect the time taken to reach the target.

Firstly, we evaluate the speed of the proposed control system. These are the error vs time curves for each scenario (averaged across all ten trials):

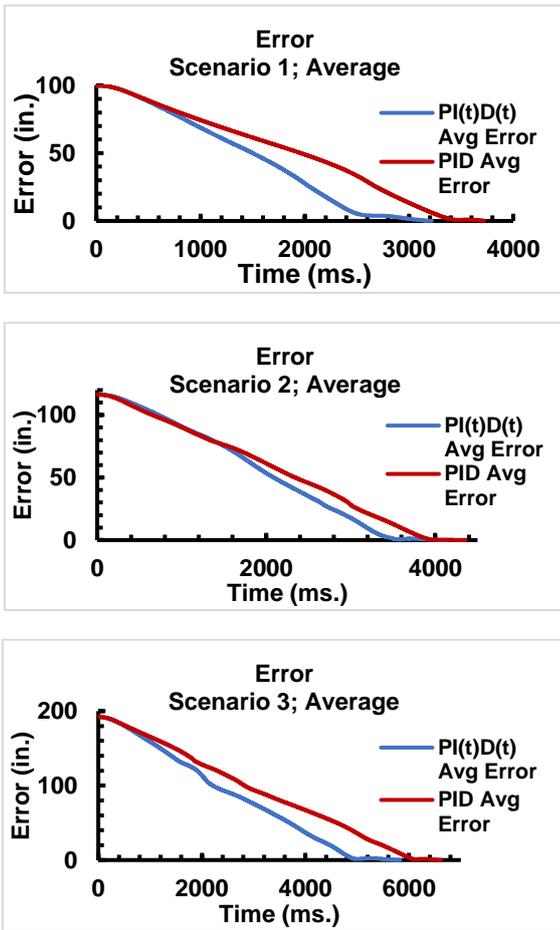

Fig. 10. Error vs Time plots for the three scenarios (red denotes PI(t)D(t) system, blue denotes PID system).

As Fig. 10. demonstrates, the PI(t)D(t) system consistently reaches the target more quickly than the PID system, despite sometimes taking more time to correct steady-state error.

The average speed of the robot for each scenario is also measured:

*Table 1: Average Speeds*

|  | Scenario 1 Avg. Speed | Scenario 2 Avg. Speed | Scenario 3 Avg. Speed |
|---|---|---|---|
| **PI(t)D(t) System** | 32.70056 in./s | 31.93082 in./s | 35.11537 in./s |
| **PID System** | 26.75437 in./s | 29.47915 in./s | 29.36590 in./s |

The average speeds corroborate these conclusions. The PI(t)D(t) system is able to consistently maintain a higher speed. The performance of PI(t)D(t) system and PID system is closest in Scenario 2, which is the path with the most simultaneous rotation. The significant amount of rotation limits the remaining power available for translation, therefore decreasing the difference in translational speed between PI(t)D(t) and PID systems. However, PI(t)D(t) system consistently outperforms the baseline.

The proposed PI(t)D(t) system demonstrates comparable safety with the PID system. The first metric for safety is path-following "error", where "error" is now taken to mean the amount of deviation from the target path (note that an acceptable amount of error is subjective to the particular application). For a visual evaluation, the real paths followed by the robot from one randomly chosen trial in each scenario (the path followed by the robot in all ten trials in the same scenario are extremely similar) is plotted on top of the target path:

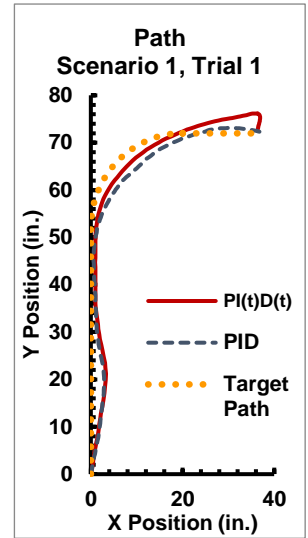

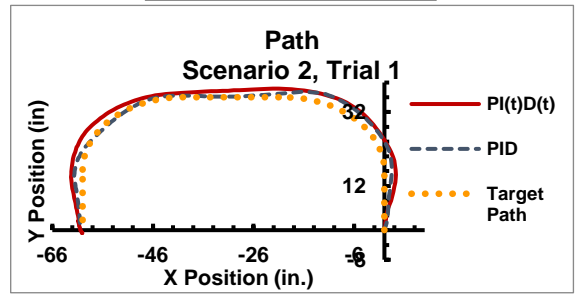

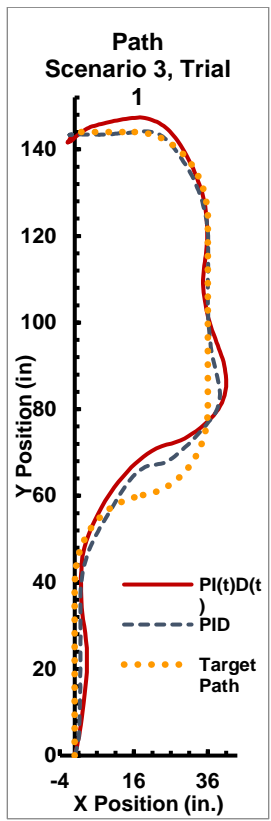

Fig. 11. Path following from each scenario.

Visually, it can be seen that the PI(t)D(t) controller generally has more error, and the error increases with increasing complexity of the path, however, the errors are not excessively significant in any scenario.

Quantatively, the average errors from the target path in each trial were also calculated. The robot's error was recorded in intervals of roughly 20 milliseconds; this interval was interpolated to 1 millisecond, and the error at each interval was averaged. The average errors from the target path for all ten trials per scenario, are listed below in *Table 2*:

*Table 2: Mean Deviation from Target Path (in.)*

|  | Trial | 1 | 2 | 3 | 4 | 5 | 6 | 7 | 8 | 9 | 10 | 10-Trial Average |
|---|---|---|---|---|---|---|---|---|---|---|---|---|
| **PI(t)D(t)** | **Scenario 1** | 1.865 | 1.777 | 1.819 | 1.769 | 1.637 | 1.933 | 1.920 | 1.733 | 1.874 | 1.908 | 1.823 |
|  | **Scenario 2** | 1.556 | 1.536 | 1.679 | 1.685 | 1.540 | 1.700 | 1.539 | 1.774 | 1.813 | 1.740 | 1.656 |
|  | **Scenario 3** | 2.379 | 2.352 | 2.332 | 2.592 | 2.365 | 2.364 | 2.650 | 2.396 | 2.563 | 2.444 | 2.444 |
| **Traditional PID** | **Scenario 1** | 1.526 | 1.610 | 1.534 | 1.532 | 1.501 | 1.566 | 1.551 | 1.581 | 1.511 | 1.599 | 1.551 |
|  | **Scenario 2** | 0.975 | 1.003 | 1.031 | 1.067 | 1.001 | 1.092 | 0.970 | 0.916 | 0.710 | 0.846 | 0.961 |
|  | **Scenario 3** | 1.506 | 1.566 | 1.526 | 1.617 | 1.520 | 1.649 | 1.642 | 1.503 | 1.564 | 1.512 | 1.561 |

The average error from the PI(t)D(t) system ranges from 18% more than the PID system for simpler paths, to 72% for paths with much more curvature. However, all the average errors remain under 2.5 inches, which is an acceptable range for a large majority of autonomous mobile robot tasks.

The maximum error, across all ten trials, for each scenario, are also listed below, in *Table 3*; this value represents the worst-case possibility when following each path.

*Table 3: Maximum Deviation from Target Path Over All Trials (in.)*

|  |  | Maximum Error |
|---|---|---|
| **PI(t)D(t)** | **Scenario 1** | 4.401 |
|  | **Scenario 2** | 3.493 |
|  | **Scenario 3** | 10.483 |
| **Traditional PID** | **Scenario 1** | 3.600 |
|  | **Scenario 2** | 2.601 |
|  | **Scenario 3** | 7.042 |

The maximum error from the PI(t)D(t) system ranges from 22% more than the PID system for simpler paths, to 49% for paths with much more curvature. For scenario 3, the deviation from the target path becomes more significant. 10.5 inches of error may be insignificant depending on the use-case and environment; however, the maximum error resulting from the PI(t)D(t) system in simpler paths, such as scenario 1 and 2, are very safe. Overall, from the perspective of deviation from the target path, PI(t)D(t) demonstrates acceptable performance.

The second metric for safety is repeatability. The paths followed by the robot in different trials of the same scenario appear nearly indistinguishable; qualitatively, the robot's motion is extremely repeatable. For a quantitative evaluation, the number of trials within various ranges of path-following error, for all three scenarios for the PI(t)D(t) system and the PID system, are plotted below in Fig. 12:

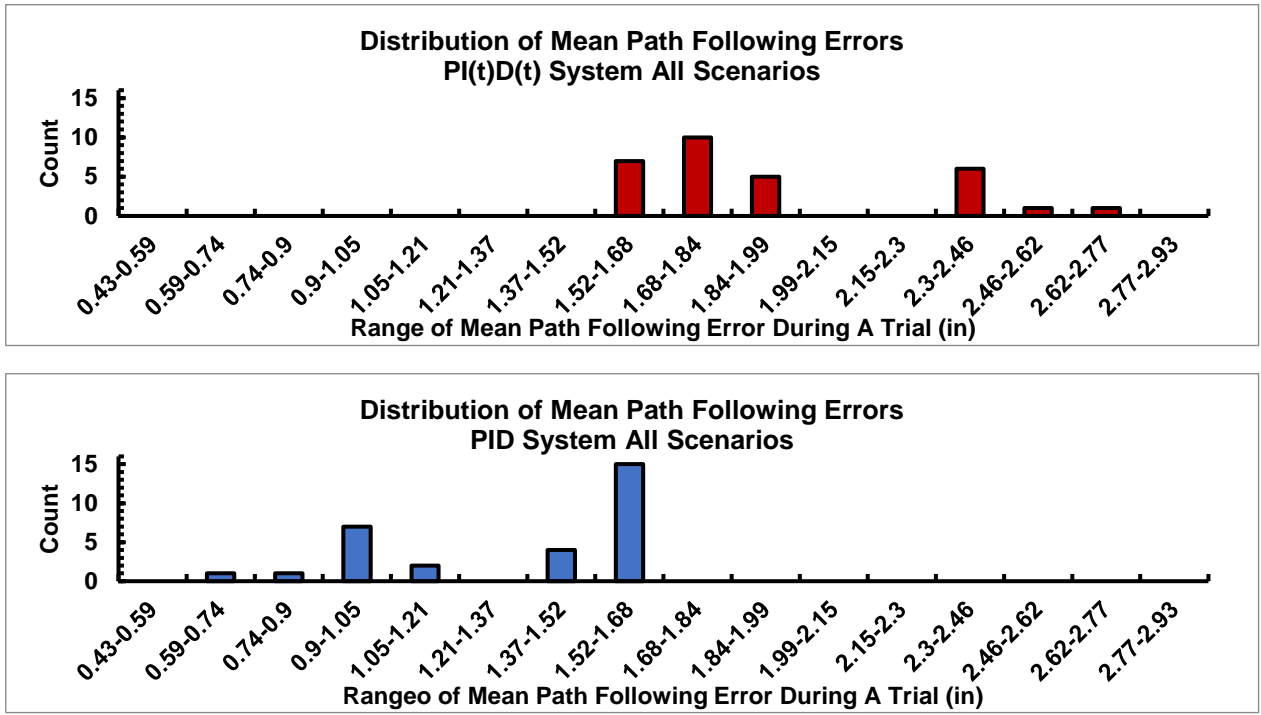

Fig. 12.　　Repeatability of Path Following Error.

The spread of error is very comparable between the PI(t)D(t) system and the PID system. For both, nearly all trials in the same scenario, and even in different scenarios, don't vary more than 1 inch in mean path following error. The repeatability in terms of path following error is very high.

The repeatability in terms of speed and timing is evaluated as well. For each scenario, the number of trials that took different ranges of time to reach the target are plotted below, in Fig. 13 (red denotes PI(t)D(t) system behavior, blue denotes PID system behavior):

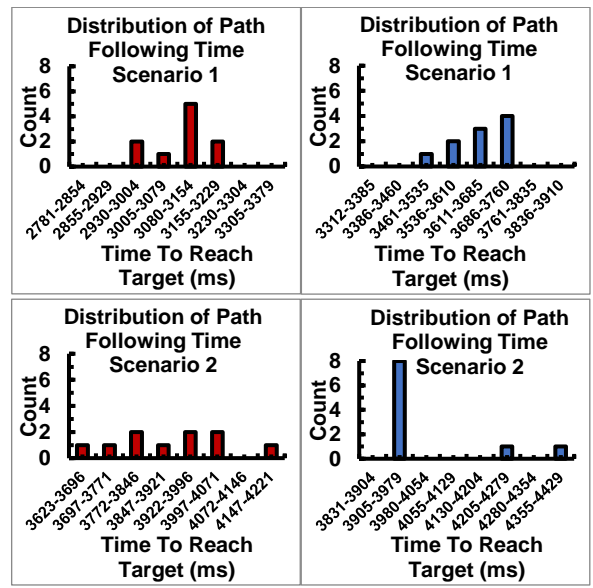

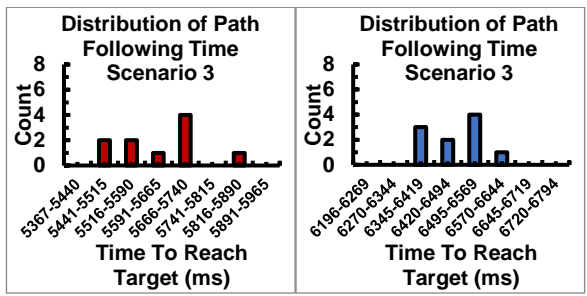

Fig. 13. Repeatability of path-following time in each trial.

(Left: PI(t)D(t) System; Right: PID System).

The spread of errors is very comparable between the PI(t)D(t) system and PID system for scenarios 1 and 3. Scenario 2 causes a larger spread of error for both the PI(t)D(t) system and PID system, likely because the high degree of simultaneous rotation in scenario 2 causes inconsistencies near the end of the path, which causes different amounts of time to be taken to correct for steady-state errors. The trend also shows that the timing inconsistency as a percentage of total path duration is likely to decrease for longer paths.

Overall, the safety of the PI(t)D(t) system is comparable to the PID system, with the PID system sometimes performing better, though both meet safety standards in most applications.

## IV. CONCLUSIONS

In this paper, a novel motion control approach is proposed. A new controller, PI(t)D(t), was developed, which scales errors for more stable behavior and more robust steady-state error correction, and sets the integral and derivative outputs as functions of time for more robust deceleration. Motion Profiling was also adapted for use in a holonomic robot, enabling smoother and more repeatable motion. Combining these elements with a pure pursuit algorithm, the overall architecture can enable greater path following speeds. Through testing a real mecanum drivebase on pre-set paths, it was determined that the PI(t)D(t) controller independently outperforms the PID controller in speed by up to 25%. In addition, the overall motion control architecture significantly outperforms a PID controller with a pure pursuit algorithm in path-following speed, and is comparable in repeatability and path-following accuracy in most cases. There are some limitations to the conducted tests; for example, the test platform is not an exact representation of real service robots because of its weight and size, and the robot was only tested on a solid flat surface. Overall, the methods described demonstrate significant potential to improve speed and efficiency of mobile robot motion, potentially enabling proliferated usage in human-servicing applications.

## REFERENCES

[1] M. B. Alatise and G. P. Hancke, "A Review on Challenges of Autonomous Mobile Robot and Sensor Fusion Methods," *IEEE Access*, vol. 8, pp. 39830-39846, February 2020.

[2] P. Raja and S.Pugazhenthi, "Optimal path planning of mobile robots: A review," *International Journal of Physical Sciences*, vol. 7(9), pp. 1314-1320, February 2012.

[3] F. Rubio, F. Valero and C. Llopis-Albert, "A review of mobile robots: Concepts, methods, theoretical framework, and applications," *International Journal of Advanced Robotic Systems,* March-April 2019.

[4] K. Berns and S. A. Mehdi, "Use of an Autonomous Mobile Robot for Elderly Care," *Advanced Technologies for Enhancing Quality of Life (AT-EQUAL)*, pp. 121-126, July 2010.

[5] J. M. Evans, "HelpMate: an autonomous mobile robot courier for hospitals," *Proceedings of IEEE/RSJ International Conference on Intelligent Robots and Systems (IROS'94)*, vol. 3, pp. 1695-1700, 1994.

[6] M. Takahashi, T. Suzuki, H. Shitamoto, T. Moriguchi, and K. Yoshida, "Developing a mobile robot for transport applications in the hospital domain," *Robotics and Autonomous Systems*, vol. 58(7), pp 889-899, July 2010.

[7] B. Burger, P. M. Maffettone, and V. V. Gusev et al, "A mobile robotic chemist," *Nature,* vol. 583, pp. 237–241, July 2020.

[8] J. Sanagate, "IDC MarketScape: Worldwide Autonomous Mobile Robots for General Warehouse Automation," [Online]. Available: https://viioni.com/wp-content/uploads/2020/02/A-IDC-AMR-Market-Report-2019.pdf

[9] M. J. Aziz Safar, "Holonomic and Omnidirectional Locomotion Systems For Wheeled Mobile Robots: A Review," *Jurnal Teknologi*, vol. 77(28), December 2015.

[10] K. J. A strom and T. H´´agglund, *PID Controllers: Theory, Design, and Tuning*, 2$^{nd}$ Ed., Research Triangle Park, NC: Instrument society of America, 1995.

[11] C. Onat, "A new design method for PI-PD control of unstable processes with dead time," *ISA Transactions,* vol. 84, pp. 69-81, January 2019.

[12] K. S. Tang, K. F. Man, G. Chen and S. Kwong, "An optimal fuzzy PID controller," *IEEE Transactions on Industrial Electronics*, vol. 48(4), pp. 757-765, 2001.

[13] Y. Mitsukura, T. Yamamoto and M. Kaneda, "A design of self-tuning PID controllers using a genetic algorithm," *Proceedings of the 1999 American Control Conference (Cat. No. 99CH36251)*, vol. 2, pp. 1361-1365.

[14] K. G. Papadopoulos, N. D. Tselepis, N. I. Margaris, "On the automatic tuning of PID type controllers via the magnitude optimum criterion," *2012 IEEE international conference on industrial technology*, 2012: pp. 869–874.

[15] H. Talebi Abatari and A. Dehghani Tafti, "Using a fuzzy PID controller for the path following of a car-like mobile robot," *2013 First RSI/ISM International Conference on Robotics and Mechatronics (ICRoM)*, 2013, pp. 189-193.

[16] D. Aneesh, "Tracking Controller of mobile robot," *2012 International Conference on Computing, Electronics and Electrical Technologies (ICCEET),* 2012, pp. 343-349.

[17] J. L. Sanchez-Lopez, P. Campoy, M. A. Olivarez Mendez, I. Mellado-Bataller and D. Galindo-Gallego, "Adaptive Control System based on Linear Control Theory for the Following Problem of a Car-Like Mobile Robot," *IFAC Proceedings Volumes*, vol. 45(3), pp. 252-257, 2012.

[18] F. J. Lin, K. K. Shyu, and C. H. Lin. "Incremental motion control of linear synchronous motor," *IEEE Transactions on Aerospace and Electronic Systems*, vol. 38(3), pp. 1011–1022, August 2002.

[19] C. Lewin, "Motion control gets gradually better," *Machine Design*, vol. 66(21), pp. 90-94, November 1994.

[20] P. H. Meckl and W. P. Seering, "Minimizing residual vibration for point-to-point motion," *Journal of Vibration, Acoustics, Stress, and Reliability in Design*, vol. 107(4), pp. 378-382, October 1994.

[21] A. Segovia, M. Rombaut, A. Preciado and D. Meizel, "Comparative study of the different methods of path generation for a mobile robot in a free environment," *Fifth International Conference on Advanced Robotics 'Robots in Unstructured Environments*, 1991, vol. 2, pp. 1667-1670.

[22] K. D. Nguyen, T. Ng, and I. Chen, "On Algorithms for Planning S-Curve Motion Profiles," *International Journal of Advanced Robotic Systems,* vol. 5(1), January 2008.

[23] R. C. Coutler, "Implementation of the Pure Pursuit Path Tracking Algorithm," The Robotics Institute, Carnegie Mellon University, Pittsburgh, Pennsylvania, January 1992.

[24] M. Samuel, M. Hussein, and M. B. Mohamad, "A Review of some Pure-Pursuit based Path Tracking Techniques for Control of Autonomous Vehicle," *International Journal of Computer Applications*, vol. 135(1), pp. 35-38, February 2016.